\documentclass[sigconf,screen,nonacm]{acmart}

\usepackage{multirow}
\usepackage{enumitem}
\usepackage{xcolor}

\definecolor{mygreen}{RGB}{0,140,72}
\definecolor{myred}{RGB}{200,55,45}
\definecolor{mygray}{RGB}{120,120,120}

\newcommand{\gup}[1]{\textcolor{mygreen}{\scriptsize$\uparrow$#1}}
\newcommand{\gdown}[1]{\textcolor{myred}{\scriptsize$\downarrow$#1}}
\newcommand{\gsame}{\textcolor{mygray}{\scriptsize$\pm$0.00}}
\AtBeginDocument{%
  }

\begin{document}

\renewcommand\footnotetextcopyrightpermission[1]{}
\title[STEAR: Layer-Aware Spatiotemporal Evidence Intervention for Hallucination Mitigation in Video-LLMs]{STEAR: Layer-Aware Spatiotemporal Evidence Intervention for Hallucination Mitigation in Video Large Language Models}


\author{Linfeng Fan}
\email{2023200424@ruc.edu.cn}
\affiliation{%
  \institution{Gaoling School of Artificial Intelligence, Renmin University of China}
  \city{Beijing}
  \country{China}
}

\author{Yuan Tian}
\email{tianyuan2004@ruc.edu.cn}
\affiliation{%
  \institution{Gaoling School of Artificial Intelligence, Renmin University of China}
  \city{Beijing}
  \country{China}
}

\author{Ziwei Li}
\email{ziwei.li@kaust.edu.sa}
\affiliation{%
  \institution{King Abdullah University of Science and Technology}
  \city{Thuwal}
  \country{Saudi Arabia}
}

\author{Zhiwu Lu}
\authornote{Corresponding author.}
\email{luzhiwu@ruc.edu.cn}
\affiliation{%
  \institution{Gaoling School of Artificial Intelligence, Renmin University of China}
  \city{Beijing}
  \country{China}
}

\renewcommand{\shortauthors}{Fan et al.}

\begin{abstract}
Video Large Language Models (Video-LLMs) remain prone to spatiotemporal hallucinations, often generating visually unsupported details or incorrect temporal relations. Existing mitigation methods typically treat hallucination as a uniform decoding failure, applying globally shared correction rules. We instead observe that decoder layers contribute differently to visual grounding and later linguistic composition, indicating that intervention must be layer-aware. Based on this insight, we propose STEAR, a layer-aware spatiotemporal evidence intervention framework. STEAR identifies high-risk decoding steps and selects token-conditioned visual evidence from grounding-sensitive middle layers. It uses this shared evidence for two coupled purposes: restoring missing local grounding in middle layers, and constructing temporally perturbed patch-level counterfactuals to falsify inconsistent reasoning during late-layer decoding. Consequently, STEAR mitigates both spatial and temporal hallucinations within an efficient single-encode inference framework. Experiments across representative Video-LLM backbones and challenging benchmarks demonstrate that STEAR consistently reduces hallucinations while improving faithfulness, temporal consistency, and robustness. Our results confirm that reliable video decoding relies on intervening on precise evidence at the right layer, rather than enforcing a global penalty. The code is provided in the Supplementary Material.
\end{abstract}

\ccsdesc[500]{Computing methodologies~Computer vision}
\settopmatter{printacmref=false} 

\keywords{Video large language models, hallucination mitigation, spatiotemporal grounding, counterfactual decoding, video understanding, multimodal generation}



\maketitle

\begin{figure*}[t!]
    \centering 
    \includegraphics[width=0.99\textwidth]{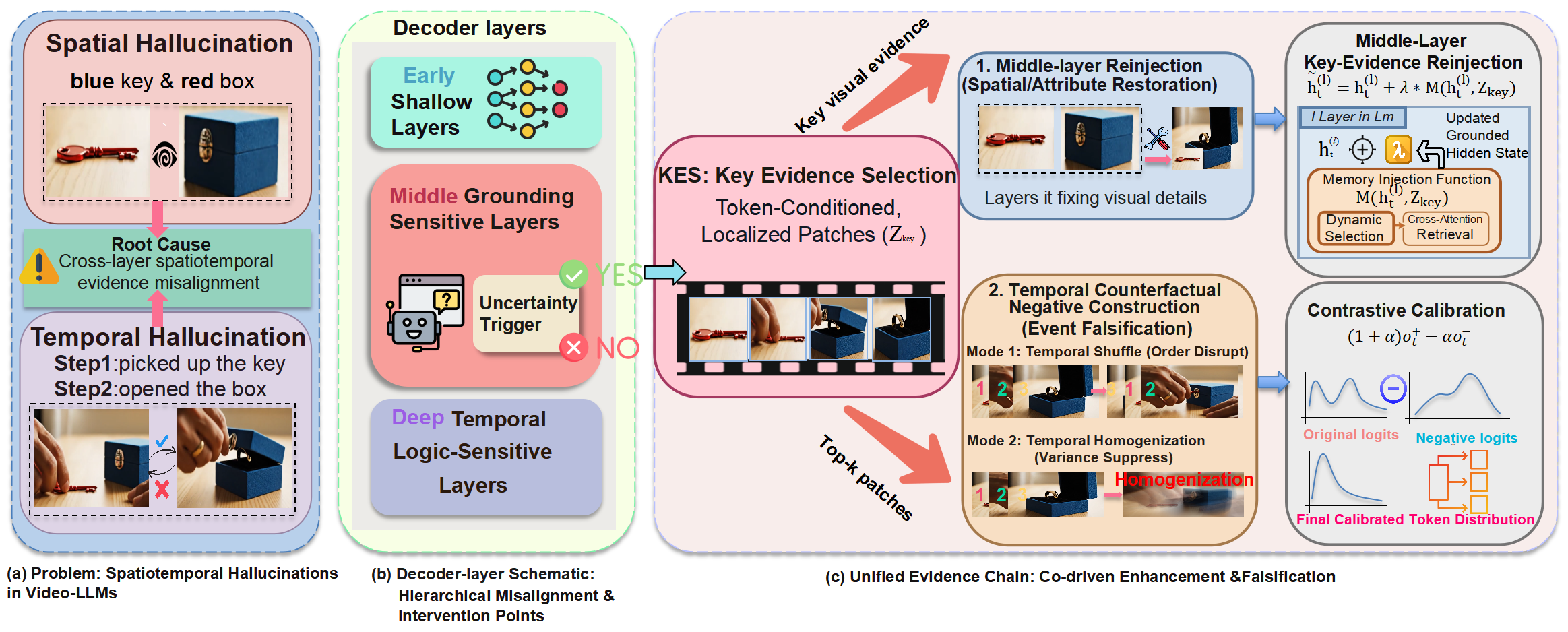}
    \caption{Given a single video encoding, STEAR first diagnoses risky decoding steps via token uncertainty, then selects token-conditioned key evidence shared by both middle-layer reinjection and deep-layer counterfactual decoding. This unified design addresses spatial hallucination by restoring missing grounding and suppresses temporal hallucination by falsifying temporally corrupted key evidence.}
    \Description{dsp1}
    \label{fig:wide1}
\end{figure*}

\section{Introduction}

 Recent advances in large language models and vision-language models \cite{radford2021learning} have substantially broadened the scope of multimodal understanding. From image-level reasoning to long-form video analysis, researchers have increasingly extended language models to richer visual environments \cite{li2023blip, liu2023visual, maaz2024video, lin2024video, xu2024pllava, cheng2024videollama,dai2023instructblip}. Among these efforts, Video Large Language Models (Video-LLMs) have become a major direction for the multimedia community, as they must jointly model appearance, motion, and event evolution, and have shown promising performance on video question answering \cite{xiao2021next, song2024moviechat,li2024mvbench}, video captioning \cite{maaz2024video, lin2024video}, and video reasoning tasks \cite{huang2024vtimellm}.

Despite these capabilities, hallucination fundamentally bottlenecks reliable \emph{video-grounded language decoding}. Unlike static factual errors in image models \cite{li2023evaluating}, video hallucinations compound across dimensions: models invent unsupported objects, misrepresent event orders, and distort cross-frame causal dependencies \cite{liu2023visual, maaz2024video, huang2024vtimellm}. These errors resist easy control; once generation drifts from valid visual evidence, language priors rapidly amplify the deviation. For practical multimedia systems, such spatiotemporal hallucination is a central reliability failure, rather than a minor artifact \cite{gao2024aigcs}.

Current mitigation strategies primarily rely on external verification, retrieval augmentation, contrastive decoding, or memory enhancement \cite{yin2024woodpecker, wang2024mementos, zou2024look, leng2024mitigating}. These methods largely share a restrictive assumption: hallucination is treated as a \emph{single-source decoding error} correctable via a uniform rule applied globally at decoding time. In video, however, hallucinations stem from diverse failures, such as missing local evidence or mistaken temporal reasoning. Crucially, decoder layers do not contribute equally to these failures. Middle layers actively align visual evidence with evolving token states, whereas deeper layers dominate linguistic composition and higher-level reasoning. Treating the decoder as a monolithic whole ignores this asymmetry, rendering intervention coarse and often unstable \cite{conmy2304towards,lin2026text,mir2025geometry}.

We observe that during video-conditioned decoding, cross-attention in middle layers exhibits more concentrated, temporally precise evidence alignment than early or late layers \cite{neo2024towards, jiang2025devils}. This indicates that Video-LLM hallucination is more accurately formulated as \emph{cross-layer spatiotemporal evidence misalignment}. Errors occur either when local evidence is insufficiently grounded in middle layers, or when late layers proceed under false temporal assumptions. Consequently, hallucination mitigation must be \emph{layer-aware}, explicitly targeting the visual evidence leveraged by the current token rather than enforcing identical corrections everywhere.

Based on this view, we propose \textbf{STEAR}, a layer-aware spatiotemporal evidence intervention framework. As illustrated in Figure~\ref{fig:wide1}, STEAR performs inference-time mitigation through a unified evidence chain. Using token uncertainty, it first isolates hallucination-prone decoding steps. It then executes \emph{Key Evidence Selection} (KES) to localize the patch-level visual evidence most relevant to the current token. Rather than treating evidence enhancement and contrastive calibration as independent modules, STEAR applies this single token-conditioned hypothesis to two coupled interventions. It reinjects the selected evidence into grounding-sensitive middle layers to restore local grounding, while simultaneously constructing temporally perturbed counterfactuals to enable late-layer contrastive decoding. Consequently, spatial errors are mitigated by repairing missing grounding, and temporal errors are suppressed by falsifying inconsistent pseudo-evidence. Crucially, all operations execute over a single video encoding pass, avoiding the latency overhead of repeated visual re-encoding.

This architecture ensures the intervention is both \emph{selective} and \emph{internally consistent}. STEAR activates exclusively at high-risk decoding steps, preventing indiscriminate token perturbation. Furthermore, because the negative branch derives directly from the positive branch's own evidence hypothesis—rather than globally corrupted views—contrastive decoding ceases to be a generic calibration tool. Instead, it becomes an explicit test of whether a prediction holds when its underlying temporal structure is falsified.

Experiments across multiple reasoning and hallucination benchmarks demonstrate that STEAR consistently reduces spatial and temporal errors. It achieves state-of-the-art or highly competitive performance on representative Video-LLMs, visibly improving faithfulness, temporal consistency, and robustness. \noindent Our contributions are summarized as follows:
\begin{itemize}[leftmargin=1.5em,itemsep=0.25em,topsep=0.25em]
\item We formulate Video-LLM hallucination as \textbf{\emph{cross-layer spatiotemporal evidence misalignment}}, demonstrating that decoder middle layers are the critical locus for precise evidence repair.
\item We present \textbf{STEAR}, a unified inference-time framework that integrates uncertainty-triggered key evidence selection, middle-layer evidence reinjection, and later-layer temporal counterfactual decoding under a shared token-conditioned evidence hypothesis.
\item We demonstrate empirically that STEAR systematically reduces spatial and temporal hallucinations, achieving \textbf{\emph{state-of-the-art}} results on challenging benchmarks while preserving the efficiency of a single-encode pipeline.
\end{itemize}
\section{Related Work}

\subsection{Video-LLMs and Video Hallucination}
Large language models have rapidly expanded from image-based reasoning to video-grounded understanding, giving rise to Video-LLMs that jointly model appearance, motion, and event evolution \cite{maaz2024video, lin2024video, cheng2024videollama, xu2024pllava, li2025videochat}. These models have shown promising performance on video question answering, captioning, and reasoning tasks, but their reliability remains constrained by hallucination \cite{kalai2025language}. Compared with image hallucination, video hallucination is intrinsically more challenging: errors may involve not only unsupported objects or attributes, but also mistaken event order, temporal transitions, and causal relations across frames. Recent benchmarks have made this issue increasingly explicit, especially for temporally misleading and event-relation-sensitive settings, such as VidHalluc and VERHallu \cite{li2025vidhalluc, zhang2026verhallu}. Our work is motivated by this setting and focuses on mitigating hallucination during video-grounded language decoding.

\subsection{Inference-time Mitigation via Grounding and Decoding}

A broad line of work mitigates multimodal hallucinations at inference time through external verification, retrieval augmentation, attention correction, memory enhancement, or contrastive decoding \cite{yin2024woodpecker, huang2024opera,li2023contrastive, zou2024look, jung2025avcd,wang2024mitigating}. In video settings, recent methods have also emphasized temporally grounded reasoning and event-aware correction \cite{song2024moviechat, huang2024vtimellm, sun2026smartsight,cai2025mitigating,su2025activation,zhou2025alw}. Among the closest decoding-based approaches, TriCD \cite{xing2026learning} introduces a plug-and-play contrastive framework for video hallucination mitigation, but relies on multiple encoding channels, auxiliary saliency or motion estimators, and tool-dependent perturbation design \cite{xing2026learning}. Event-relation methods such as KFP, evaluated on VERHallu \cite{zhang2026verhallu}, show that redistributing attention around key frames can improve event-level factuality without retraining \cite{zhang2026verhallu}. These methods provide an important foundation for video-time hallucination mitigation.

Our work is related to, but distinct from, this line. Memory-based approaches can recover visual support, but usually do not explicitly model where in the decoder grounding is most effectively repaired. Decoding-based approaches can suppress language prior, but typically operate on the final output distribution or on globally modified views. In contrast, STEAR is a training-free, backbone-agnostic inference-time method that uses a \emph{single-encode}, \emph{layer-aware}, and \emph{shared-evidence} design: it first restores token-conditioned local evidence in grounding-sensitive middle layers, and then falsifies the same evidence hypothesis through localized patch-level temporal counterfactual decoding.

\subsection{Layer-wise Evidence Routing and Intervention}
Recent analyses of multimodal decoding and mechanistic interpretability \cite{zou2023representation,clark2019does} suggest that decoder layers play different functional roles in grounding and language generation. Works such as PlaM provide evidence for layer-wise specialization, where early layers separate signals, middle layers align multimodal evidence, and later layers become increasingly language-dominant \cite{bachu2024layer, neo2024towards, jiang2025devils,tao2024probing}. Related findings further indicate that aggregating cross-layer attention can yield more stable evidence estimates, as explored by CLAA\cite{mcdanel2026claa} and related attention-aggregation analyses\cite{meng2022locating}. These observations are highly relevant to video-conditioned decoding, yet remain underexploited in most hallucination mitigation methods.

Our work builds directly on this emerging perspective. STEAR uses middle-layer attention not as a generic saliency cue, but as a stable routing signal for token-conditioned Key Evidence Selection (KES), which is then shared by both grounding repair and temporal falsification. In this sense, STEAR is closer in spirit to layer-aware intervention than to monolithic decoding calibration.

\section{Method}

\subsection{Overview and Layer-wise Diagnostics}

We address hallucinations in Video-LLMs as a problem of \emph{cross-layer spatiotemporal evidence misalignment}. The issue is not merely that a token is predicted incorrectly at one step; rather, the model fails to invoke the right visual evidence at the right stage of decoding. When token-relevant evidence is not sufficiently consolidated in grounding-sensitive layers, later layers tend to continue decoding under incomplete or distorted support, which eventually manifests as spatial or temporal hallucinations.

\begin{figure}[t!]
    \centering
    \includegraphics[width=\linewidth]{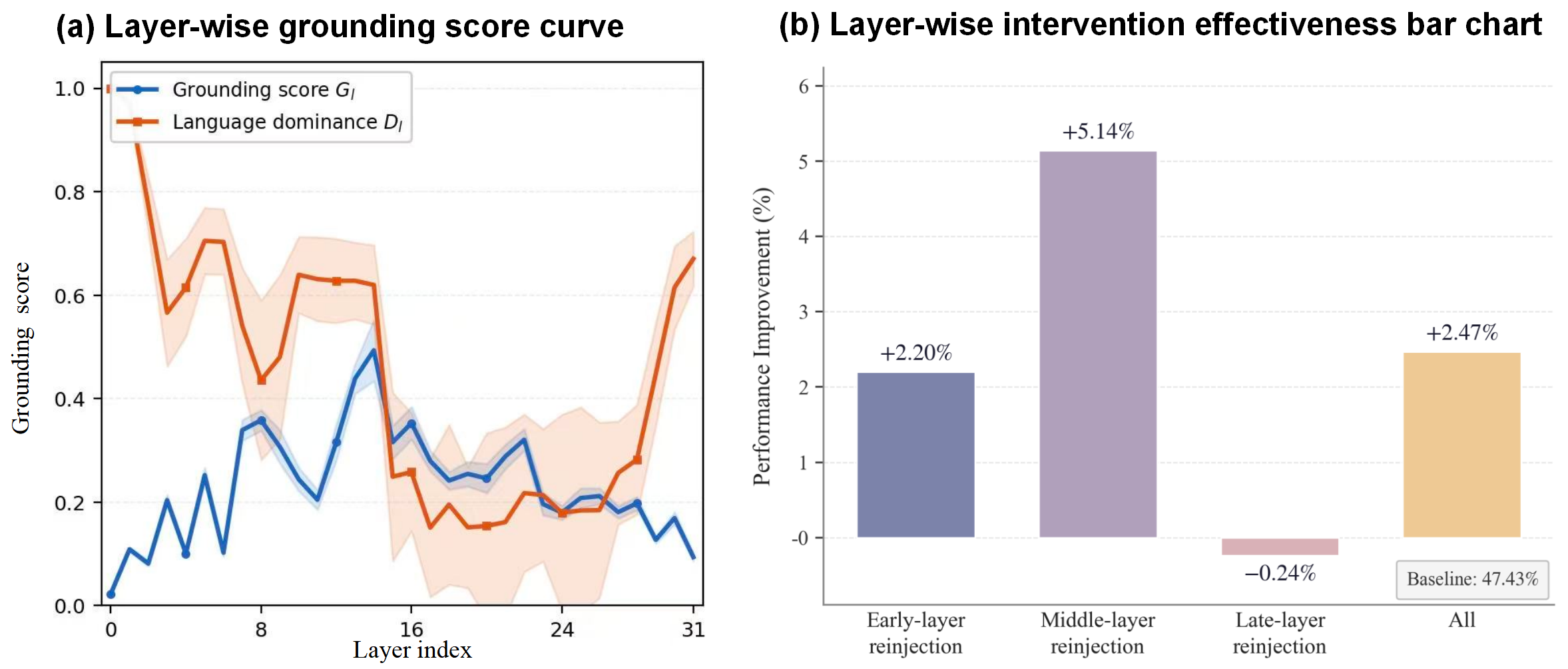}
    \caption{(a) Grounding score peaks in the middle decoder layers, while language dominance drops in the middle and rises again in the late layers, indicating a transition from evidence consolidation to language-dominant decoding. This suggests that hallucinations arise when token-relevant visual evidence is not sufficiently grounded before the model enters the late reasoning regime.
(b) Middle-layer reinjection yields the largest gain, whereas others are ineffective.}
    \Description{dsp2}
    \label{fig:wide2}
\end{figure}

To motivate where intervention should occur, we first perform a layer-wise diagnostic analysis, summarized in Fig.~\ref{fig:wide2}. For a token $t$ and decoder layer $l$, let $\mathbf{A}_t^{(l)} \in \mathbb{R}^{N}$ denote the cross-attention weights from the current token to visual tokens, and let
{\setlength{\abovedisplayskip}{3pt}
\setlength{\belowdisplayskip}{3pt}
\begin{equation}
\mathbf{p}_t^{(l)}=\operatorname{softmax}\!\bigl(\mathbf{W}_{\mathrm{lm}}\operatorname{LN}(\mathbf{h}_t^{(l)})\bigr)
\end{equation}}
be the per-layer token distribution obtained by reusing the model's original final layer norm and LM head, without introducing any auxiliary classifier.

We define the \textbf{grounding score} at layer $l$ based on benchmark annotations rather than self-selected top-attention patches. Specifically, let $\mathcal{E}_t$ denote the annotated evidence region for token $t$ in the benchmark, after projecting the annotation onto the visual patch-token space. We compute
{\setlength{\abovedisplayskip}{0pt}
\setlength{\belowdisplayskip}{0pt}\begin{equation}
G(l)=\mathbb{E}_{t}\!\left[\sum_{i\in \mathcal{E}_t} A_{t,i}^{(l)}\right].
\end{equation}}
A higher $G(l)$ indicates that a larger fraction of the layer's cross-attention is assigned to benchmark-annotated evidence relevant to the current token. To quantify how strongly a layer is dominated by text prior rather than visual input, we further define the \textbf{language-dominance score}
{\setlength{\abovedisplayskip}{2pt}
\setlength{\belowdisplayskip}{2pt}
\begin{equation}
D(l)=\mathbb{E}_{t}\!\left[\cos\!\left(\mathbf{p}_t^{(l)}, \mathbf{p}_{t,\varnothing}^{(l)}\right)\right],
\end{equation}}
where $\mathbf{p}_{t,\varnothing}^{(l)}$ is the per-layer token distribution obtained from the same decoder state under masked visual input.\footnote{This masked-input forward pass is used only for diagnosis in Fig.~\ref{fig:wide2}, not during inference.}
A higher $D(l)$ indicates stronger agreement with a text-only decoding tendency. Fig.~\ref{fig:wide2}(a) shows that benchmark-annotated evidence alignment peaks in the middle decoder layers, while language dominance is relatively lower there and rises again in the late layers. Fig.~\ref{fig:wide2}(b) further shows that intervention is empirically most effective in the same middle region. Together, these diagnostics motivate a layer-aware design in which evidence is \emph{read} and \emph{repaired} in the middle layers before late-layer language priors dominate the prediction.

\begin{figure*}[t!]
    \centering 
    \includegraphics[width=\textwidth]{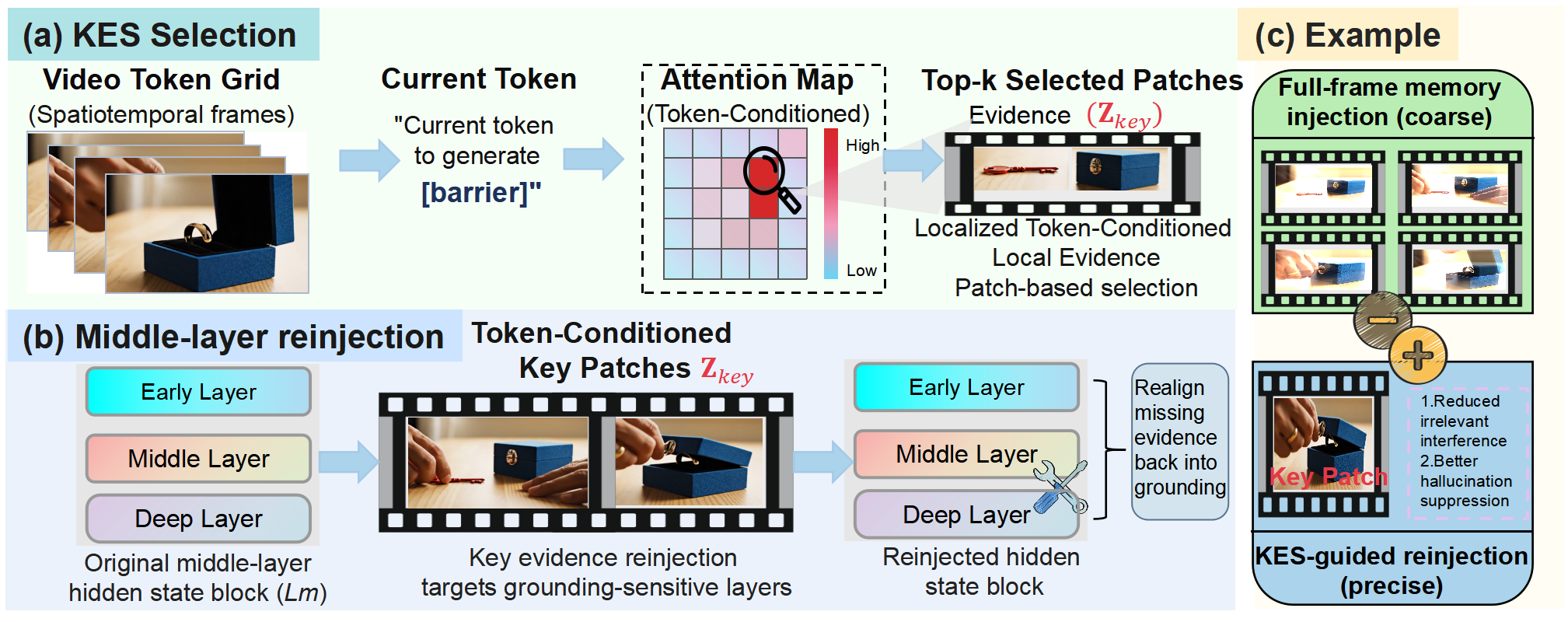}
    \caption{For each risky token, KES aggregates middle-layer cross-attention to identify the top-$k$ key patches that best support the current prediction. The selected evidence is then reinjected into grounding-sensitive middle layers, restoring token-required but potentially neglected visual support without disturbing stable decoding steps.}
    \Description{dsp2}
    \label{fig:wide3}
\end{figure*}

Based on this view, we propose \textbf{STEAR}, a layer-aware spatiotemporal evidence intervention framework for Video-LLMs. STEAR is built on one unifying principle: \textbf{the same token-conditioned evidence hypothesis should govern both positive grounding enhancement and negative counterfactual suppression}. Concretely, STEAR first identifies risky decoding steps by token uncertainty. It then performs \textbf{Key Evidence Selection} (KES) from middle-layer cross-attention to localize the patch-level spatiotemporal evidence most relevant to the current token. As illustrated in Fig.~\ref{fig:wide3}, the selected evidence is reinjected into grounding-sensitive middle layers in a MemVR-style \cite{zou2024look} FFN retracing manner to restore missing visual support. At the same time, the same evidence is temporally perturbed to construct a counterfactual negative branch for later-layer contrastive decoding. All operations are performed on top of a single video encoding pass.

\subsection{Uncertainty-Triggered Key Evidence Selection}

Given a video $V=\{f_1,\dots,f_T\}$ and a text prompt $Q$, a Video-LLM encodes the video into visual tokens $\mathbf{Z}=\{z_1,\dots,z_N\}$ and autoregressively predicts the next token under
{\setlength{\abovedisplayskip}{2pt}
\setlength{\belowdisplayskip}{2pt}
\begin{equation}
p(y_t \mid y_{<t}, \mathbf{Z}).
\end{equation}}
Let $\mathbf{h}_t^{(l)}$ denote the hidden state at decoding step $t$ and decoder layer $l$. Since hallucinations are sparse and concentrated on a small subset of tokens, STEAR is activated only at genuinely risky positions.

We monitor token uncertainty within a predefined set of middle layers $\mathcal{L}_m$ and define the risk score as
{\setlength{\abovedisplayskip}{3pt}
\setlength{\belowdisplayskip}{3pt}
\begin{equation}
u_t=\max_{l\in\mathcal{L}_m}\mathcal{U}(\mathbf{p}_t^{(l)}),
\end{equation}}
where $\mathcal{U}(\cdot)$ is instantiated as normalized entropy:
{\setlength{\abovedisplayskip}{3pt}
\setlength{\belowdisplayskip}{3pt}
\begin{equation}
\mathcal{U}(\mathbf{p})=-\frac{1}{\log |\mathcal{V}|}\sum_{j=1}^{|\mathcal{V}|} p_j \log p_j,
\end{equation}}
with $|\mathcal{V}|$ denoting the vocabulary size. When $u_t>\tau$, the current token is identified as high-risk and STEAR is triggered. We define the trigger layer as
{\setlength{\abovedisplayskip}{0pt}
\setlength{\belowdisplayskip}{3pt}
\begin{equation}
l^{\ast}=\arg\max_{l\in\mathcal{L}_m}\mathcal{U}(\mathbf{p}_t^{(l)}),
\end{equation}}
which specifies the decoder depth from which the counterfactual branch is launched.

Once triggered, we perform \textbf{Key Evidence Selection (KES)}. The goal of KES is not to retrieve an entire frame or a global memory bank, but to estimate which \emph{local spatiotemporal evidence} the current token truly depends on. Motivated by Fig.~\ref{fig:wide2}, KES is derived from middle-layer cross-attention, since this region exhibits the strongest alignment with benchmark-annotated evidence and the most stable grounding behavior. Let $\mathcal{N}(l^{\ast})$ denote a small middle-layer neighborhood around the trigger layer. We aggregate cross-attention over this neighborhood:
{\setlength{\abovedisplayskip}{3pt}
\setlength{\belowdisplayskip}{3pt}
\begin{equation}
\mathbf{a}_t=\frac{1}{|\mathcal{N}(l^{\ast})|}\sum_{l\in\mathcal{N}(l^{\ast})}\mathbf{A}_t^{(l)}.
\end{equation}}
We then select the top-$k$ most relevant visual patches:
{\setlength{\abovedisplayskip}{3pt}
\setlength{\belowdisplayskip}{3pt}
\begin{equation}
\mathcal{I}_t=\operatorname{TopK}(\mathbf{a}_t,r),
\end{equation}}
where $r$ is the patch selection ratio. The corresponding evidence subset is denoted by $\mathbf{Z}_{\text{key}}$.

We do not claim that cross-attention is a perfect attribution signal. In STEAR, it is used as a \emph{routing signal} for intervention. To reduce head-level and layer-level noise, KES aggregates attention across a grounding-sensitive middle-layer neighborhood, and its effectiveness is later validated by random-selection, frame-level, and decoupled-selector ablations. In this way, KES serves as the shared evidence hypothesis of the whole framework: it determines both what should be restored in the positive branch and what should be challenged in the negative branch.

\subsection{Middle-Layer Key-Evidence Reinjection}

After obtaining the key evidence $\mathbf{Z}_{\text{key}}$, we do not attach it to the final output layer. Instead, we reinject it into \emph{grounding-sensitive} middle-layer representations, as illustrated in Fig.~\ref{fig:wide3}. The rationale is that middle layers are the primary locus where visual evidence is aligned with evolving token states. If crucial local evidence is ignored or gradually attenuated at this stage, later layers can only continue decoding on top of incomplete evidence, making hallucination much more likely \cite{li2024multi}.

We treat the selected visual evidence as supplementary \emph{key-value memory} and reinject it through the FFN at the middle trigger region, rather than fusing it only at the output side. Importantly, STEAR is \emph{training-free}: it introduces no additional training data, loss, or parameter updates. Specifically, $\mathbf{W}_v$ reuses the backbone's original visual projection into the decoder space, while the remaining reinjection tensors are constructed on the fly at inference time rather than learned as new trainable parameters. Thus, the reported STEAR results already correspond to the setting \emph{without any learnable additions trained for STEAR}.

Concretely, given the selected patch set $\mathcal{I}_t$, we first collect the corresponding dimension-aligned visual memory entries
{\setlength{\abovedisplayskip}{2pt} \setlength{\belowdisplayskip}{2pt} 
\begin{equation}
\mathbf{M}_t=\{\mathbf{W}_v z_i \mid i\in\mathcal{I}_t\},
\end{equation}}
where each $\mathbf{m}_{t,j}\in\mathbf{M}_t$ is a frozen projected visual memory entry. For each layer $l\in\mathcal{N}(l^{\ast})$, we retrieve from this selected memory using the current hidden state as the query:
{\setlength{\abovedisplayskip}{2pt} \setlength{\belowdisplayskip}{2pt} 
\begin{equation}
\beta_{t,j}^{(l)}
=
\frac{
\exp\!\bigl(
\phi_q^{(l)}(\operatorname{LN}(\mathbf{h}_t^{(l)}))^\top
\phi_k^{(l)}(\mathbf{m}_{t,j})
\bigr)
}{
\sum_{j'}
\exp\!\bigl(
\phi_q^{(l)}(\operatorname{LN}(\mathbf{h}_t^{(l)}))^\top
\phi_k^{(l)}(\mathbf{m}_{t,j'})
\bigr)
},
\end{equation}}
and compute the retrieved evidence as
{\setlength{\abovedisplayskip}{3pt} \setlength{\belowdisplayskip}{3pt} 
\begin{equation}
\mathbf{r}_t^{(l)}
=
\mathbf{W}_o^{(l)}
\sum_j
\beta_{t,j}^{(l)}\,\phi_v^{(l)}(\mathbf{m}_{t,j}).
\end{equation}}
We then inject this retrieved signal through the FFN path:
{\setlength{\abovedisplayskip}{2pt} \setlength{\belowdisplayskip}{1pt} 
\begin{equation}
\tilde{\mathbf{h}}_t^{(l)}
=
\mathbf{h}_t^{(l)}+\lambda\,\mathbf{r}_t^{(l)},
\quad l\in\mathcal{N}(l^{\ast}).
\end{equation}}

In this way, the selected patch-level evidence is not simply appended as an extra feature, but reintroduced into the decoder precisely where grounding repair is most effective. Compared with directly using global visual features, this design is more targeted: it refreshes the token-required local evidence that may have been weakened or neglected during decoding, while preserving the layer-aware intervention principle of STEAR. In practice, this step mainly targets spatial hallucinations caused by missing objects, local attributes, and misjudged spatial relations.

\subsection{Patch-level Temporal Counterfactual Decoding}

Positive evidence reinjection alone is still insufficient for temporal hallucination. Many failures occur not because the model fails to perceive a relevant region \cite{jiang2024prior}, but because it organizes that region under an incorrect temporal or causal relation. To address this issue, we further introduce \textbf{patch-level temporal counterfactual decoding}, which falsifies the same key evidence hypothesis used by the positive branch \cite{huang2024neighbor}.

Specifically, we reuse the key patch set $\mathcal{I}_t$ selected by KES and perturb only the temporal dimension of these patches to construct the negative visual evidence $\mathbf{Z}^{-}$. We consider two complementary perturbation modes. The first is temporal shuffle,
{\setlength{\abovedisplayskip}{2pt} \setlength{\belowdisplayskip}{2pt} \begin{equation}
z^{-}_{i,1:T}=\operatorname{Shuffle}(z_{i,1:T}), \quad i\in\mathcal{I}_t,
\end{equation}}
and the second is temporal homogenization,
{\setlength{\abovedisplayskip}{0pt} \setlength{\belowdisplayskip}{0pt} \begin{equation}
z^{-}_{i,\tau}=(1-\gamma)z_{i,\tau}+\gamma\bar{z}_i,\quad
\bar{z}_i=\frac{1}{T}\sum_{\tau=1}^{T}z_{i,\tau}, \quad i\in\mathcal{I}_t.
\end{equation}}
The former explicitly destroys order information, while the latter suppresses temporal variation. Together, they create a localized counterfactual that preserves most spatial semantics while corrupting the temporal structure most relevant to the current token.

This perturbation is applied in visual-token space after video encoding. As a result, the temporal slot structure and positional scaffold of the backbone remain intact, while only the content assigned to the selected temporal slots is modified. In other words, STEAR does not test arbitrary noise injection; it tests whether the prediction remains stable when temporally critical content is made inconsistent with its original temporal support. This distinction is important because it limits the distribution shift introduced by the negative branch and better matches the goal of temporal falsification.

We then branch directly from the trigger layer instead of re-encoding the whole video:
{\setlength{\abovedisplayskip}{3pt} \setlength{\belowdisplayskip}{3pt} \begin{equation}
\mathbf{H}^{-}_{t,L}
=
\mathcal{F}_{L:l^{\ast}+1}\bigl(\mathbf{H}^{-}_{t,l^{\ast}},\mathbf{Z}^{-}\bigr),
\end{equation}}
where $\mathcal{F}_{L:l^{\ast}+1}$ denotes the original decoder from the trigger layer to the final layer. Let $\mathbf{o}_t^{+}$ and $\mathbf{o}_t^{-}$ denote the logits of the positive and negative branches. The final calibrated logits are computed as
{\setlength{\abovedisplayskip}{3pt} \setlength{\belowdisplayskip}{3pt} \begin{equation}
\mathbf{o}_t=(1+\alpha)\mathbf{o}_t^{+}-\alpha \mathbf{o}_t^{-}.
\end{equation}}

The key point is not merely that we apply contrastive decoding, but that we perform \textbf{evidence-level falsification}. The positive branch assumes that $\mathbf{Z}_{\text{key}}$ provides the crucial support for the current prediction, while the negative branch tests this same assumption under temporal corruption. If a candidate token remains highly probable even after the temporal structure of the key evidence has been broken, then that token is more likely supported by language prior than by genuine video evidence and should be suppressed. In this way, STEAR turns contrastive decoding from a generic distribution calibration technique into an explicit test of whether the current prediction is still valid under temporally corrupted evidence.

Overall, STEAR forms a unified inference-time evidence chain: uncertainty decides when to intervene, KES decides which evidence matters, middle-layer reinjection restores grounding for that evidence, and later-layer counterfactual decoding falsifies the same evidence hypothesis under temporal corruption. For low-risk tokens, decoding proceeds normally. For high-risk tokens, STEAR performs only local enhancement and local falsification where needed, thereby achieving a favorable balance among intervention strength, inference cost, and practical robustness.

\begin{table*}[t]
\caption{Cross-backbone overall comparison. All numbers are accuracies (\%). Avg. denotes the arithmetic mean over the four reported overall metrics under the same backbone. For each non-baseline method, the inline delta shows the absolute change relative to the corresponding baseline of the same backbone. Best and second-best results within each backbone are highlighted in bold and underline, respectively.}
\label{tab:main_overall}
\centering
\small
\scalebox{0.98}{
\tabcolsep14pt
\begin{tabular}{llccccc}
\toprule
Backbone & Method & EventHallusion$\uparrow$ & VidHalluc$\uparrow$ & NExT-QA$\uparrow$ & MVBench$\uparrow$ & Avg.$\uparrow$ \\
\midrule
\multirow{6}{*}{LLaVA-Video-7B}
& Baseline
& 46.70 {\gsame}
& 63.07 {\gsame}
& 58.43 {\gsame}
& \underline{41.28} {\gsame}
& 52.37 {\gsame} \\
& +VCD
& 40.17 {\gdown{6.53}}
& 65.48 {\gup{2.41}}
& 60.27 {\gup{1.84}}
& 38.92 {\gdown{2.36}}
& 51.21 {\gdown{1.16}} \\
& +MemVR
& 47.10 {\gup{0.40}}
& \underline{65.69} {\gup{2.62}}
& \underline{60.47} {\gup{2.04}}
& 41.25 {\gdown{0.03}}
& \underline{53.63} {\gup{1.26}} \\
& +TCD
& \underline{51.62} {\gup{4.92}}
& 64.39 {\gup{1.32}}
& 60.27 {\gup{1.84}}
& 37.47 {\gdown{3.81}}
& 53.44 {\gup{1.07}} \\
& +DINO-HEAL
& 46.70 {\gsame}
& 65.58 {\gup{2.51}}
& 60.27 {\gup{1.84}}
& 40.94 {\gdown{0.34}}
& 53.37 {\gup{1.00}} \\
& +\textbf{STEAR}(Ours)
& \textbf{54.77} {\gup{8.07}}
& \textbf{65.81} {\gup{2.74}}
& \textbf{61.88} {\gup{3.45}}
& \textbf{41.74} {\gup{0.46}}
& \textbf{56.05} {\gup{3.68}} \\
\cmidrule(lr){1-7}

\multirow{6}{*}{InternVL2.5-8B}
& Baseline
& 69.44 {\gsame}
& 70.20 {\gsame}
& 78.76 {\gsame}
& \underline{63.21} {\gsame}
& 70.40 {\gsame} \\
& +VCD
& \textbf{73.84} {\gup{4.40}}
& 68.47 {\gdown{1.73}}
& 79.64 {\gup{0.88}}
& 63.11 {\gdown{0.10}}
& 71.27 {\gup{0.87}} \\
& +MemVR
& 67.48 {\gdown{1.96}}
& 70.69 {\gup{0.49}}
& \underline{82.03} {\gup{3.27}}
& 63.11 {\gdown{0.10}}
& 70.83 {\gup{0.43}} \\
& +TCD
& 70.66 {\gup{1.22}}
& \underline{71.51} {\gup{1.31}}
& 79.52 {\gup{0.76}}
& 62.74 {\gdown{0.47}}
& 71.11 {\gup{0.71}} \\
& +DINO-HEAL
& 70.28 {\gup{0.84}}
& 71.09 {\gup{0.89}}
& 82.11 {\gup{3.35}}
& 63.18 {\gdown{0.03}}
& \underline{71.67} {\gup{1.27}} \\
& +\textbf{STEAR}(Ours)
& \underline{72.37} {\gup{2.93}}
& \textbf{73.19} {\gup{2.99}}
& \textbf{82.62} {\gup{3.86}}
& \textbf{68.21} {\gup{5.00}}
& \textbf{74.10} {\gup{3.70}} \\
\cmidrule(lr){1-7}

\multirow{6}{*}{Qwen2.5-VL-7B}
& Baseline
& 63.81 {\gsame}
& 81.70 {\gsame}
& 59.63 {\gsame}
& 58.74 {\gsame}
& 65.97 {\gsame} \\
& +VCD
& 64.06 {\gup{0.25}}
& \underline{83.19} {\gup{1.49}}
& 60.55 {\gup{0.92}}
& \underline{61.45} {\gup{2.71}}
& \underline{67.31} {\gup{1.34}} \\
& +MemVR
& 64.06 {\gup{0.25}}
& 81.87 {\gup{0.17}}
& \underline{60.77} {\gup{1.14}}
& 59.50 {\gup{0.76}}
& 66.55 {\gup{0.58}} \\
& +TCD
& 62.35 {\gdown{1.46}}
& 82.67 {\gup{0.97}}
& 60.43 {\gup{0.80}}
& 60.82 {\gup{2.08}}
& 66.57 {\gup{0.60}} \\
& +DINO-HEAL
& \underline{64.79} {\gup{0.98}}
& 82.95 {\gup{1.25}}
& 60.13 {\gup{0.50}}
& 59.76 {\gup{1.02}}
& 66.91 {\gup{0.94}} \\
& +\textbf{STEAR}(Ours)
& \textbf{67.48} {\gup{3.67}}
& \textbf{83.84} {\gup{2.14}}
& \textbf{60.87} {\gup{1.24}}
& \textbf{61.58} {\gup{2.84}}
& \textbf{68.44} {\gup{2.47}} \\
\bottomrule
\end{tabular}}
\vspace{0.05in}
\end{table*}

\section{Experiments}

\subsection{Experimental Protocol}

We evaluate STEAR on four representative benchmarks that together cover both hallucination-sensitive and general video understanding settings: \textbf{EventHallusion} \cite{zhang2024eventhallusion}, which focuses on event-level hallucination under misleading evidence; \textbf{VidHalluc} \cite{li2025vidhalluc}, which diagnoses temporal hallucination under multiple question formats; \textbf{NExT-QA} \cite{xiao2021next}, which emphasizes causal, temporal, and descriptive reasoning; and \textbf{MVBench} \cite{li2024mvbench}, which measures general video understanding across diverse fine-grained tasks. Following prior work on multimodal hallucination mitigation \cite{leng2024mitigating, zou2024look, li2025vidhalluc, zhang2024eventhallusion}, we report the official accuracy-based metrics of each benchmark, where higher is better.

We conduct experiments on three representative Video-LLM backbones: \textbf{LLaVA-Video-7B} , \textbf{InternVL2.5-8B}  and \textbf{Qwen2.5-VL-7B} \cite{zhang2024llava,chen2024expanding,bai2025qwen25vltechnicalreport, bai2025qwen3}. These models span different grounding capacities and decoding behaviors, allowing us to examine whether STEAR generalizes across architectures rather than overfitting to a single model family. We compare STEAR with representative inference-time mitigation methods, including \textbf{VCD}, \textbf{MemVR}, \textbf{TCD}, and \textbf{DINO-HEAL} \cite{leng2024mitigating, zou2024look, li2025vidhalluc, zhang2024eventhallusion}. Unless otherwise specified, STEAR monitors token uncertainty within decoder layers 5--16 and triggers intervention when the uncertainty score exceeds $\tau=0.85$. For Key Evidence Selection (KES), we select the top $10\%$ attention-salient patches. The temporal homogenization coefficient is set to $\gamma=0.80$, and the contrastive strength is set to $\alpha=0.75$. The negative branch is executed without gradient computation and reuses the original visual encoding. All other decoding settings follow the default configurations of the corresponding backbones.

Since complete cross-backbone runs are currently available for EventHallusion \cite{zhang2024eventhallusion}, NExT-QA  \cite{xiao2021next}, and MVBench  \cite{li2024mvbench}, the main comparison in this section focuses on these three benchmarks. Detailed VidHalluc \cite{li2025vidhalluc} results are reported in the subsequent analysis section, where they are used to validate the temporal robustness of STEAR more directly. Due to space constraints, additional comparisons with recent inference-time mitigation methods, parameter-sensitivity studies and cross-scale generalization experiments are reported in the supplementary material.

\subsection{Main Comparison Across Backbones}

Table~\ref{tab:main_overall} reports the overall comparison across three backbones on four benchmarks. On \textbf{LLaVA-Video-7B} \cite{zhang2024llava}, STEAR achieves the best performance on all four metrics, improving EventHallusion \cite{zhang2024eventhallusion} from 46.70 to 54.77, VidHalluc \cite{li2025vidhalluc} from 63.07 to 65.81, NExT-QA \cite{xiao2021next} from 58.43 to 61.88, and MVBench \cite{li2024mvbench} from 41.28 to 41.74. The largest gain appears on EventHallusion, where STEAR surpasses the strongest baseline TCD\cite{zhang2024eventhallusion} by 3.15 points, indicating that layer-aware evidence intervention is particularly effective when hallucination is severe.

On the stronger \textbf{InternVL2.5-8B} \cite{chen2024expanding} backbone, STEAR delivers the best average result and the strongest performance on three of the four benchmarks, including VidHalluc, NExT-QA, and MVBench. Although VCD remains slightly better on EventHallusion, STEAR provides the most balanced improvement across hallucination-sensitive and general reasoning settings. This pattern is consistent with our motivation: when the backbone is already better grounded, generic contrastive calibration can remain competitive on a single benchmark, whereas STEAR more reliably improves the broader evidence-use behavior of the decoder.

On \textbf{Qwen2.5-VL-7B} \cite{bai2025qwen25vltechnicalreport, bai2025qwen3}, STEAR again achieves the best result on all four reported metrics. The consistency across three distinct model families suggests that the gain of STEAR is not tied to a particular architecture, but arises from its underlying principle of layer-aware evidence intervention.

\begin{table*}[t!]
\caption{Fine-grained breakdown on LLaVA-Video-7B. STEAR is particularly strong on misleading event cases and consistently improves all NExT-QA and MVBench sub-groups. Best results are in bold and second-best results are underlined.}
\label{tab:fine_grained_llava}
\centering
\small
\scalebox{0.98}{
\tabcolsep6.5pt
\begin{tabular}{lccccccccccccc}
\toprule
Method
& \multicolumn{4}{c}{EventHallusion$\uparrow$}
& \multicolumn{4}{c}{NExT-QA$\uparrow$}
& \multicolumn{5}{c}{MVBench$\uparrow$} \\
\cmidrule(lr){2-5}\cmidrule(lr){6-9}\cmidrule(lr){10-14}
& Entire & Misleading & Mix & Overall
& C & T & D & Acc.
& Loc. & Pred. & Seq. & Obj. & Overall \\
\midrule
Baseline
& 29.82 & \underline{43.13} & 58.55 & 46.70
& 57.32 & 53.44 & 72.53 & 58.43
& \underline{29.33} & 36.60 & \underline{55.12} & \underline{41.60} & \underline{41.28} \\

+MemVR
& 30.70 & 42.15 & \underline{59.06} & 47.10
& \underline{59.65} & \underline{55.33} & \underline{73.87} & \underline{60.47}
& 28.83 & \underline{36.80} & \underline{55.12} & 41.30 & 41.25 \\

+VCD
& 28.90 & 43.10 & 45.10 & 40.10
& 59.38 & \underline{55.33} & 73.49 & 60.27
& 29.17 & 33.10 & 52.00 & 39.60 & 38.92 \\

+TCD
& \textbf{47.36} & \underline{43.13} & 58.54 & \underline{51.62}
& 59.61 & 55.02 & 73.36 & 60.27
& 27.33 & 33.60 & 47.00 & 39.20 & 37.47 \\

+DINO-HEAL
& 28.90 & 43.10 & \textbf{59.10} & 46.70
& 59.49 & 55.09 & 73.62 & 60.27
& \underline{29.33} & 35.70 & 54.75 & 41.30 & 40.94 \\
\midrule
+\textbf{STEAR (Ours)}
& \underline{42.11} & \textbf{73.53} & 52.33 & \textbf{54.77}
& \textbf{60.88} & \textbf{56.88} & \textbf{74.48} & \textbf{61.88}
& \textbf{30.50} & \textbf{37.00} & \textbf{55.62} & \textbf{42.10} & \textbf{41.74} \\
\bottomrule
\end{tabular}}
\vspace{0.05in}
\end{table*}

\subsection{Fine-grained Results and Efficiency}

To understand where the gains come from, Table~\ref{tab:fine_grained_llava} reports a fine-grained breakdown on \textbf{LLaVA-Video-7B} \cite{zhang2024llava}, where the grounding gap is largest and the effect of inference-time intervention is the clearest. Three observations are especially important.

First, on \textbf{EventHallusion} \cite{zhang2024eventhallusion}, STEAR shows a very large gain on the \emph{Misleading} split, improving from 43.13 to 73.53. This is the most telling fine-grained result in the paper. It suggests that STEAR is not merely improving easier cases, but is particularly effective when the model is exposed to plausible yet misleading evidence, exactly where spatiotemporal hallucination is hardest to control.

Second, on \textbf{NExT-QA} \cite{xiao2021next}, STEAR improves all three reasoning categories, including causal, temporal, and descriptive questions. This indicates that the gain is not limited to a single subtype of reasoning, but comes from better evidence use during decoding.

Third, on \textbf{MVBench} \cite{li2024mvbench}, STEAR improves all reported task groups and achieves the best overall score. The improvement is modest but consistent, which is important: hallucination mitigation is obtained without sacrificing general video understanding ability.

A practical advantage of STEAR is shown in Table~\ref{tab:efficiency}. Unlike conventional contrastive decoding methods that require separately encoded counterfactual views \cite{zhang2024eventhallusion,jung2025avcd}, STEAR preserves a \emph{single-encode} inference pipeline by reusing the original visual tokens and launching the negative branch only after the trigger layer. As a result, it achieves a better efficiency--effectiveness trade-off than prior contrastive decoding baselines: lower relative latency, lower memory overhead, and stronger EventHallusion performance. This suggests that the gain of STEAR does not come from brute-force extra computation, but from intervening at more informative layers and on more relevant evidence.

\begin{table}[t!]
\caption{Efficiency comparison on LLaVA-Video-7B. ``Enc.'' denotes the number of visual encoding passes, and ``EH'' refers to EventHallusion. STEAR preserves a single-encode design while achieving the strongest hallucination mitigation performance.}
\label{tab:efficiency}
\centering
\small
\setlength{\tabcolsep}{5pt}
\resizebox{\columnwidth}{!}{
\begin{tabular}{lcccc}
\toprule
Method & Enc. & Rel. Latency$\downarrow$ & Rel. Memory$\downarrow$ & EH Overall$\uparrow$ \\
\midrule
VCD           & 2 & 2.42$\times$ & 1.13$\times$ & 40.17 \\
TCD           & 2 & 2.25$\times$ & 1.13$\times$ & 51.62 \\
STEAR (Ours)  & 1 & \textbf{1.89$\times$} & \textbf{1.01$\times$} & \textbf{54.77} \\
\bottomrule
\end{tabular}
}
\end{table}

\section{Insight Validation and Analysis}

This section tests whether the mechanism claimed by STEAR is actually supported by evidence, rather than merely reporting additional ablations. Our central claim is that hallucination in Video-LLMs is better understood as \emph{cross-layer spatiotemporal evidence misalignment}: token-relevant evidence is not sufficiently preserved in the middle layers, and later layers may further amplify the error under temporal or causal priors. Accordingly, Table~\ref{tab:ablation_core} is organized around three questions: whether intervention should be selective, why the middle layers are the right locus for evidence repair, and why local temporal counterfactuals are more effective than coarse perturbations.

\begin{table}[t!]
\caption{Core ablations on LLaVA-Video-7B for validating the mechanism of STEAR. The table is organized to test three claims: selective intervention via uncertainty triggering, grounding repair in middle layers, and evidence-level temporal falsification through local counterfactuals. Inline deltas show absolute changes relative to the baseline.}
\label{tab:ablation_core}
\centering
\small
\scalebox{0.98}{
\tabcolsep3pt
\begin{tabular}{p{0.57\columnwidth}cc}
\toprule
Variant & \shortstack[c]{EventHallusion\\Overall$\uparrow$} & \shortstack[c]{VidHalluc\\(ACH-BQA)$\uparrow$} \\
\midrule
Baseline & 47.43 {\gsame} & 41.02 {\gsame} \\
\textbf{STEAR (full)} & \textbf{54.77} {\gup{7.34}} & \textbf{44.14} {\gup{3.12}} \\

\midrule
\multicolumn{3}{l}{\textit{Core components}} \\
\shortstack[l]{w/o uncertainty trigger\\(apply to all tokens)} & 48.41 {\gup{0.98}} & 41.91 {\gup{0.89}} \\
w/o middle-layer reinjection & 50.61 {\gup{3.18}} & 43.50 {\gup{2.48}} \\
w/o temporal counterfactual decoding & 47.92 {\gup{0.49}} & 41.37 {\gup{0.35}} \\

\midrule
\multicolumn{3}{l}{\textit{Evidence selection and branch design}} \\
w/o KES (random patches) & 54.03 {\gup{6.60}} & 43.86 {\gup{2.84}} \\
\shortstack[l]{frame-level evidence selection\\instead of KES} & 53.79 {\gup{6.36}} & 43.62 {\gup{2.60}} \\
\shortstack[l]{separate selectors for positive/negative\\branches} & 52.08 {\gup{4.65}} & 43.89 {\gup{2.87}} \\

\midrule
\multicolumn{3}{l}{\textit{Intervention depth}} \\
early-layer reinjection only & 49.63 {\gup{2.20}} & 42.16 {\gup{1.14}} \\
middle-layer reinjection only & 52.57 {\gup{5.14}} & 43.85 {\gup{2.83}} \\
late-layer reinjection only & 47.19 {\gdown{0.24}} & 41.12 {\gup{0.10}} \\

\midrule
\multicolumn{3}{l}{\textit{Counterfactual construction}} \\
patch-level shuffle only & 54.01 {\gup{6.58}} & 44.10 {\gup{3.08}} \\
patch-level homogenization only & 51.34 {\gup{3.91}} & 44.02 {\gup{3.00}} \\
frame-level temporal perturbation & 52.57 {\gup{5.14}} & 43.86 {\gup{2.84}} \\
whole-video perturbation & 50.12 {\gup{2.69}} & 42.51 {\gup{1.49}} \\
\bottomrule
\end{tabular}}
\end{table}

\begin{figure*}[t!]
    \centering 
    \includegraphics[width=\textwidth]{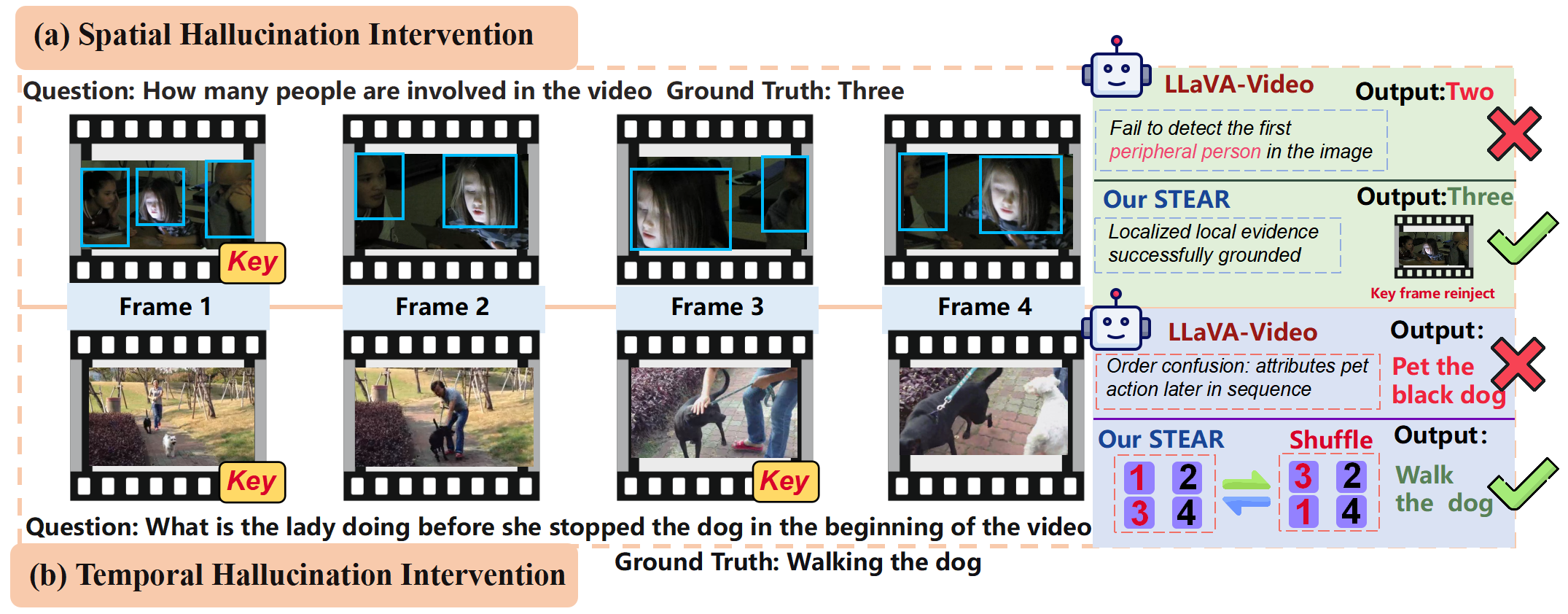}
    \caption{Compared with Video-LLMs, STEAR better grounds local evidence and suppresses temporally inconsistent generations. }
    \Description{dsp5}
    \label{fig:wide}
\end{figure*}

\subsection{Why Selective and Layer-aware Intervention?}

A first question is whether hallucination mitigation should be applied uniformly to all decoding steps, or only to genuinely risky ones. The answer is clear from Table~\ref{tab:ablation_core}. Removing the uncertainty trigger and applying STEAR to all tokens reduces EventHallusion \cite{zhang2024eventhallusion} from 54.77 to 48.41 and VidHalluc \cite{li2025vidhalluc} from 44.14 to 41.91. This shows that more intervention is not necessarily better: indiscriminate correction disrupts many already stable decoding steps and weakens the effect where intervention is truly needed.

The same table also clarifies where intervention should occur. Among single-location variants, \emph{middle-layer reinjection only} performs best, reaching 52.57 on EventHallusion and 43.85 on VidHalluc, clearly outperforming both \emph{early-layer reinjection only} (49.63 / 42.16) and \emph{late-layer reinjection only} (47.19 / 41.12). This pattern is consistent with our hypothesis. Early layers are still too close to raw visual propagation, while late layers are already strongly dominated by linguistic continuation. Middle layers provide the most effective locus because they are where token-conditioned evidence is most actively grounded. Fig.~\ref{fig:wide2} provides a direct visualization of this layer-wise tendency.


\subsection{Why Does Shared Key Evidence Selection Matter?}

A distinctive property of STEAR is that the same Key Evidence Selection (KES) mechanism is shared by both positive reinjection and negative counterfactual construction. Table~\ref{tab:ablation_core} shows that this design matters for two reasons.

First, replacing KES with random patches or frame-level evidence selection leads to a consistent drop, from 54.77 / 44.14 to 54.03 / 43.86 and 53.79 / 43.62, respectively. This indicates that token-conditioned local evidence is more informative than coarse or unguided selection.

More importantly, using \emph{separate selectors} for the positive and negative branches causes a larger degradation, reducing performance to 52.08 / 43.89. This suggests that the main value of KES is not only better localization, but also \emph{causal consistency} across branches. When the positive branch restores one evidence hypothesis while the negative branch perturbs another, contrastive decoding becomes a generic disturbance rather than an evidence-level falsification test. STEAR instead uses the same token-conditioned evidence hypothesis to decide both what should be restored and what should be challenged, which turns the framework into a coherent intervention mechanism rather than a loose combination of grounding enhancement and contrastive decoding.

\subsection{Why do Local Temporal Counterfactuals Work?}

The third question concerns the design of the negative branch. Table~\ref{tab:ablation_core} shows that temporal hallucination is exposed more effectively by \emph{local} rather than global corruption. Patch-level temporal perturbations outperform both frame-level and whole-video perturbations. In particular, frame-level temporal perturbation reaches 52.57 / 43.86, whereas whole-video perturbation drops to 50.12 / 42.51. The gap is meaningful: globally corrupted negatives introduce a larger distribution shift, but do not necessarily test the token-relevant evidence responsible for the hallucination.

Among local counterfactuals, shuffle is more effective than homogenization on EventHallusion (54.01 vs.\ 51.34), while the two are nearly comparable on VidHalluc (44.10 vs.\ 44.02). This suggests two complementary temporal failure modes: order confusion and dynamic flattening. Shuffle targets the former more directly, while homogenization captures the latter. Their combination gives the best result (54.77 / 44.14), confirming the benefit of modeling both.

The necessity of temporal counterfactual decoding is further verified by the \emph{w/o temporal counterfactual decoding} ablation, which drops sharply to 47.92 / 41.37 and approaches the baseline. This result is important because it shows that positive grounding repair alone is insufficient. Recovering token-relevant evidence helps, but it does not fully prevent later layers from organizing that evidence under an incorrect temporal or causal hypothesis. The negative branch is therefore not an auxiliary refinement; it is necessary for testing whether the current prediction remains valid under temporal falsification.

\subsection{Qualitative Analysis}

Figure~\ref{fig:wide} shows two representative cases. In the spatial case, the video contains three people at the beginning, but only two remain visible for most of the later frames. The baseline predicts \emph{two} instead of the ground-truth \emph{three}. STEAR corrects this error by recovering token-relevant local evidence from the grounding-sensitive middle layers, allowing the missing peripheral person to be properly counted. In the temporal case, the person first walks the dog, then stops, and only afterward pets the black dog. The baseline incorrectly relies on a later salient action and answers \emph{pet the black dog}. By counterfactually perturbing the key evidence along the temporal axis, STEAR suppresses this order-confused prediction and recovers the correct answer, \emph{walking the dog}. These cases are consistent with the quantitative results: STEAR improves video-grounded generation by both repairing missing local grounding and rejecting temporally inconsistent reasoning. Additional failure cases are provided in the Supplementary Material.
\section{Conclusion}

In this paper, we presented STEAR, a layer-aware intervention framework that mitigates hallucinations in Video-LLMs by reformulating them as cross-layer spatiotemporal evidence misalignments rather than uniform decoding errors. To correct these misalignments, STEAR operationalizes an efficient single-encode pipeline that integrates uncertainty-triggered key evidence selection, middle-layer evidence reinjection, and patch-level temporal counterfactual decoding. Evaluations across representative Video-LLM backbones confirm that this architecture consistently suppresses hallucinations in challenging settings. Crucially, our findings establish that reliable video-grounded decoding relies less on broad global correction than on targeted precision—specifically, intervening on the right evidence at the optimal layer and decoding step.

\bibliographystyle{ACM-Reference-Format}
\bibliography{sample-base} 

\end{document}